\documentclass{article}

\usepackage[final]{neurips_2023_ml4ps}

\usepackage{apalike}

\usepackage[utf8]{inputenc} 
\usepackage[T1]{fontenc}    
\usepackage{hyperref}       
\usepackage{url}            
\usepackage{booktabs}       
\usepackage{amsfonts}       
\usepackage{nicefrac}       
\usepackage{microtype}      
\usepackage{xcolor}         
\usepackage{graphicx}
\usepackage{pifont}
\usepackage{nicematrix}
\newcommand{\cmark}{\ding{51}}%
\newcommand{\xmark}{\ding{55}}%

\title{A Multimodal Dataset and Benchmark for Radio Galaxy and Infrared Host Detection}
\author{%
  Nikhel Gupta\thanks{nikhel.gupta@csiro.au} \\
  CSIRO Space \& Astronomy, \\
  PO Box 1130, Bentley WA 6102, Australia \\
  \And
  Zeeshan Hayder \\
  CSIRO Data61 \\
  Black Mountain ACT 2601, Australia \\
  \AND
  Ray P. Norris \\
  CSIRO Space \& Astronomy, \\
  PO Box 76, Epping, NSW 1710, Australia \\
  \And
  Minh Hyunh \\
  CSIRO Space \& Astronomy, \\
  PO Box 1130, Bentley WA 6102, Australia \\
  \And
  Lars Petersson \\
  CSIRO Data61 \\
  Black Mountain ACT 2601, Australia \\
}

\begin{document}
\maketitle

\begin{abstract}
We present a novel multimodal dataset developed by expert astronomers to automate the detection and localisation of multi-component extended radio galaxies and their corresponding infrared hosts.
The dataset comprises 4,155 instances of galaxies in 2,800 images with both radio and infrared modalities.
Each instance contains information on the extended radio galaxy class, its corresponding bounding box that encompasses all of its components, pixel-level segmentation mask, and the position of its corresponding infrared host galaxy.
Our dataset is the first publicly accessible dataset that includes images from a highly sensitive radio telescope, infrared satellite, and instance-level annotations for their identification. 
We benchmark several object detection algorithms on the dataset and propose a novel multimodal approach to identify radio galaxies and the positions of infrared hosts simultaneously. 
\end{abstract}

\section{Introduction}
\label{SEC:Intro}
Recent advancements in radio astronomy have enabled us to scan large areas of the sky in a short timescale while generating incredibly sensitive continuum images of the Universe.
This has created new possibilities for detecting millions of galaxies at radio wavelengths. 
For example, the ongoing Evolutionary Map of the Universe \citep[EMU;][]{norris21} survey, conducted using the Australian Square Kilometre Array Pathfinder \citep[ASKAP;][]{hotan21} telescope, is projected to discover more than 40 million compact and extended galaxies in the next five years \citep{norris21,hotan21}. 
Similarly, the Low-Frequency Array \citep[LOFAR;][]{vanharleem13} survey of the entire northern sky is also expected to detect more than 10 million galaxies.
With the advent of the Square Kilometre Array (SKA\footnote{https://www.skatelescope.org/the-ska-project/}) radio telescope, which is expected to become operational in the coming years, the number of galaxy detections is expected to increase further, potentially reaching hundreds of millions.
Such an enormous dataset will significantly impact our understanding of the physics of galaxy evolution. It will allow us to constrain the theoretical models of the Universe (e.g. the Big Bang model) at unprecedented levels.
To capture the full potential of these radio surveys comes the need to redesign the galaxy detection techniques.

\begin{figure}[!ht]
\centering
\includegraphics[width=12.8cm, scale=0.5]{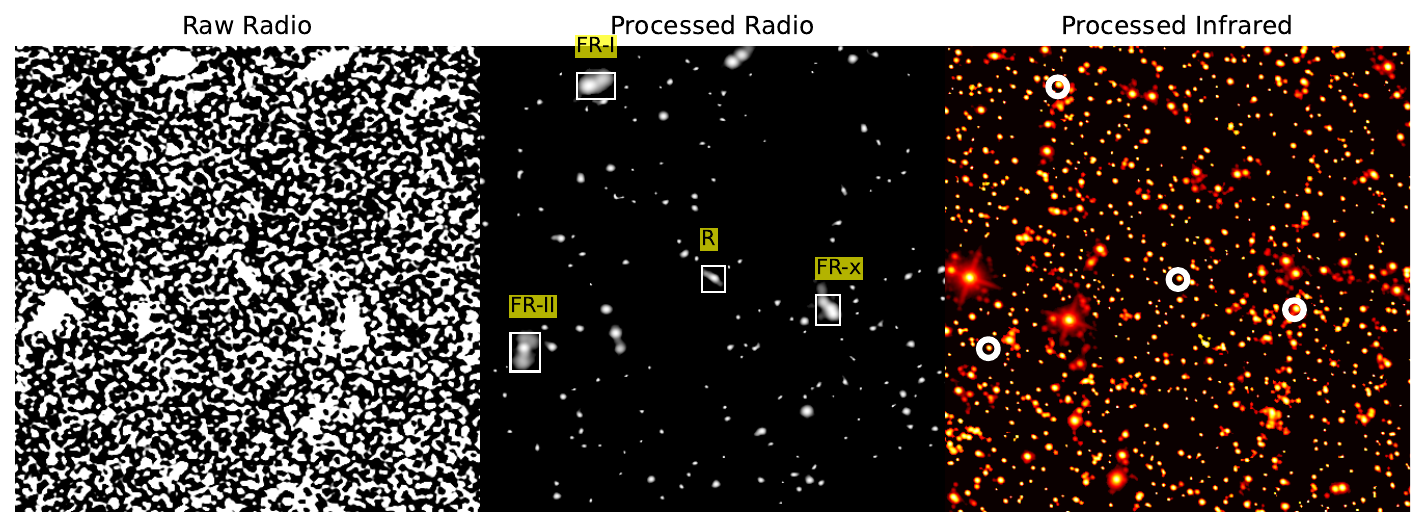}
\includegraphics[trim=0cm 0cm 0cm 1.15cm, width=12.8cm, scale=0.5]{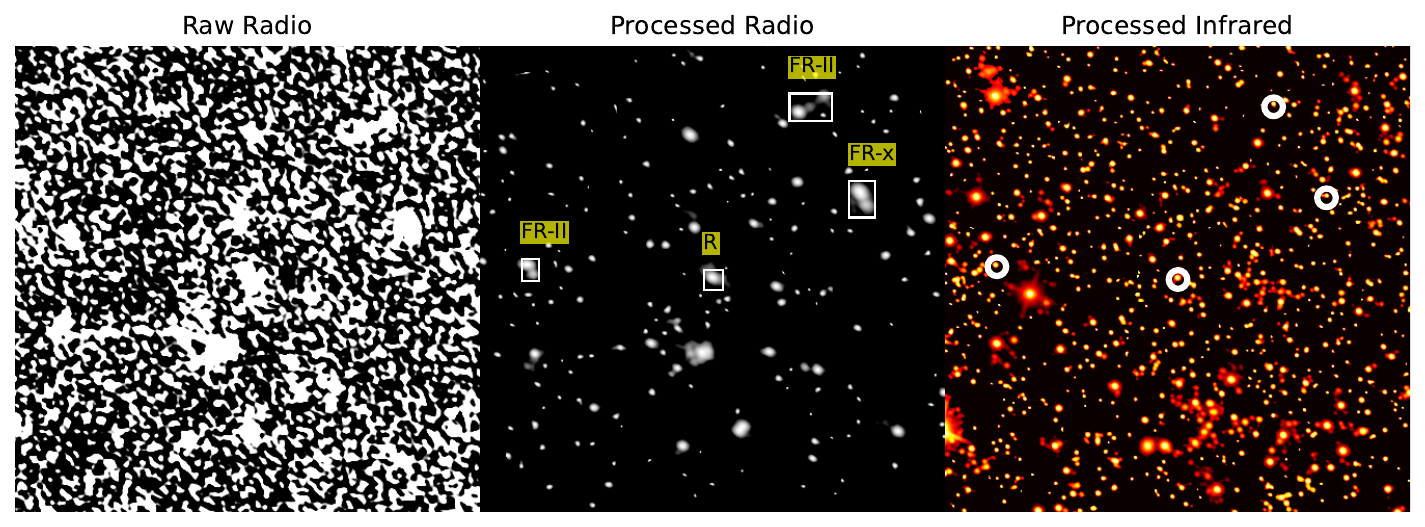}
\includegraphics[trim=0cm 0cm 0cm 1.15cm, width=12.8cm, scale=0.5]{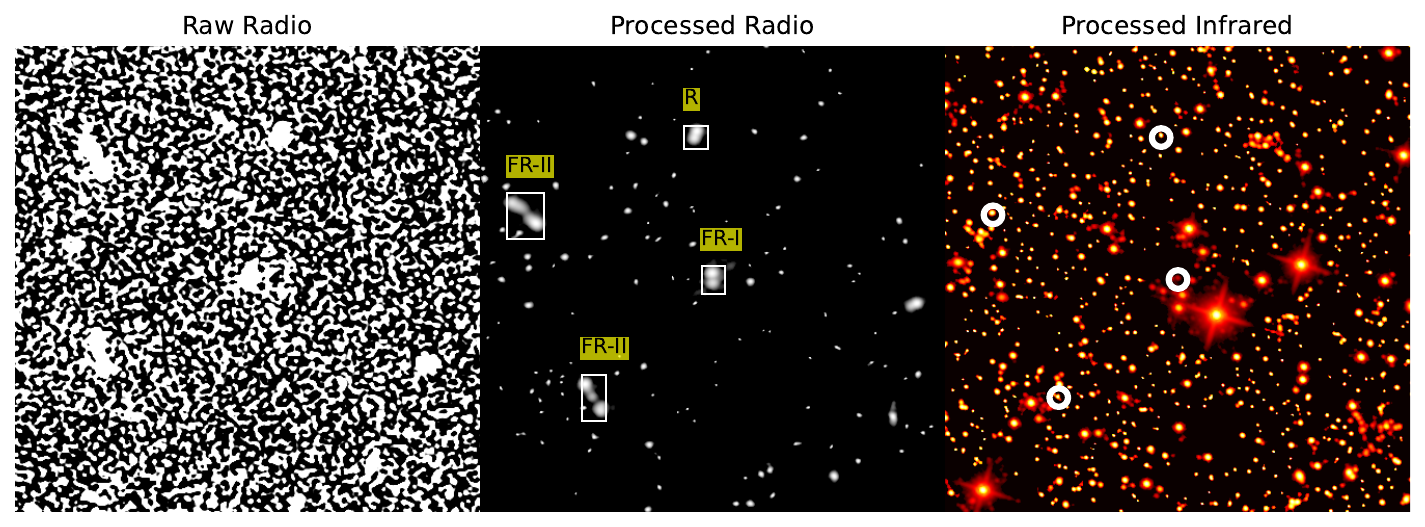}
\caption{Raw radio (left), processed radio (middle) and processed infrared (right) images with the frame size of $450\times450$ pixels ($0.25^{\circ}\times 0.25^{\circ}$). The processed radio images highlight the categories of extended radio galaxies, and the bounding boxes denote their total radio extent encompassing all of its components. The infrared images show host galaxies inside the circles.} 
\label{FIG:RadioGalaxies15arcm}
\end{figure}

Radio galaxies are characterized by giant radio emission regions that extend well beyond their structure at visible and infrared wavelengths. 
While most radio galaxies typically appear as simple, compact circular sources, increasing the sensitivity of radio telescopes result in the detection of more radio galaxies with complex extended structures. 
These structures typically consist of multiple components with distinct peak radio emissions. 
Figure~\ref{FIG:RadioGalaxies15arcm} displays examples of these extended radio galaxies in the first (raw noisy data) and second (processed data) columns, along with their compact infrared host galaxies in the third column.
To construct scientifically useful catalogues of radio galaxies, it is crucial to group the associated components of extended radio galaxies accurately. 
Currently, visual inspections are used to cross-identify associated radio source components and their infrared host galaxies.
This limitation highlights the critical need for developing automated methods, such as machine learning algorithms, to accurately and efficiently cross-identify and group associated components. 
However, to train and test such algorithms, a large and diverse dataset of labelled radio galaxy images is necessary. 
Unfortunately, such a dataset is not currently available to train models for the next generation of radio surveys, which poses a significant challenge to developing automated methods for detecting and grouping components of radio galaxies. 
This paper introduces a novel dataset aimed at addressing the problem of radio galaxy component association. 
The dataset has been structured in the COCO dataset format \citep{lin14M}, allowing for straightforward comparison studies of various object detection strategies for the machine learning community.
It features 2,800 3-channel images, each containing two radio sky channels, one corresponding infrared sky channel, and 4,155 annotations.
To summarize, our work contributes to the following aspects:
\begin{itemize}
\setlength\itemsep{0.5em}
\item We introduce the first publicly available dataset curated by professional astronomers that includes state-of-the-art images from a highly sensitive radio telescope and instance-level annotations for extended radio galaxies.
\item As a novel addition, our dataset also includes corresponding images of the infrared sky, along with the positional information of the host galaxies.
\item We benchmark the object detection algorithms on our dataset to demonstrate the challenge of detecting and associating components of radio galaxies. Additionally, we propose a novel method to detect the positions of infrared host galaxies simultaneously.
\end{itemize}

\section{The Dataset}
\label{SEC:dataset}
\subsection{Radio and Infrared Images}
Our dataset contains radio images derived from observations with the ASKAP telescope. 
We use the Evolutionary Map of Universe pilot survey \citep[EMU-PS;][]{norris21} that covers a sky area of 270 deg$^2$, achieving an RMS sensitivity of $25-35~\mu$Jy/beam at a frequency range of 800 to 1088 MHz, centred at 944 MHz (wavelength of 0.37 to 0.28m, centred at 0.32m). 
The extended radio galaxies were visually identified by the experts in the 270 deg$^2$ EMU-PS image.
At the same sky locations of radio images, we obtain AllWISE \citep[][]{cutri13} infrared images from the Wide-field Infrared Survey Explorer's \citep[WISE;][]{wright10} W1 band that correspond to 3.4 $\mu$m wavelength.
We create 3-channel RGB images by combining the processed radio and infrared images.
To achieve this, we fill the B and G channels with 8-16 bit and 0-8 bit radio information, respectively. 
In contrast, the 8-16 bit infrared information is inserted into the R channel.

\subsection{Annotations}
\begin{figure}[!ht]
\centering
\includegraphics[width=13.5cm, scale=0.5]{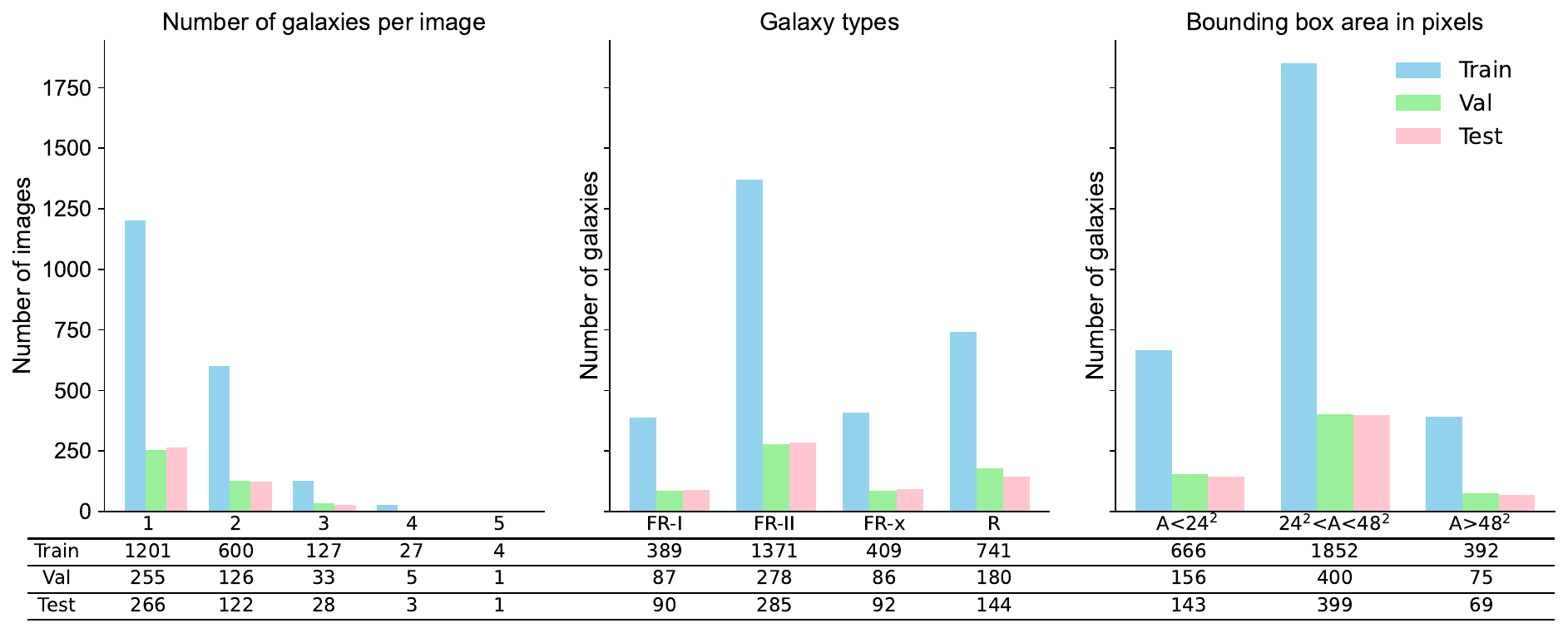}
\caption{The dataset split distributions. Shown are the distributions of extended radio galaxies in one frame (left), their categories (middle) and the occupied area per galaxy.}
\label{FIG:Statistics}
\end{figure}

Our dataset comprises four types of annotations: the classification labels for extended radio galaxies, bounding boxes encompassing all components of each radio galaxy, segmentation masks for radio galaxies, and the positions of infrared host galaxies.
The comprehensive methodology for source identification will be presented in detail by \citet{yew23prep}. 
Here, we provide a brief overview of the process.
We visually inspected infrared images to determine the infrared host galaxy associated with each radio source.
Following the criteria of \citet{fanaroff74}, we classified the galaxies as FR-I and FR-II.
The unreliable classifications, which can either be FR-I or FR-II in reality, are labelled as FR-x sources.
In some cases, barely resolved sources have only one peak outside the central component, we classify them as R (for "resolved") sources. 
The radio annotations for each galaxy are stored as `categories', `bbox', and `segmentation'. 
The positions of the infrared hosts are stored as `keypoints'. 
The statistics for the train, validation, and test data splits, including the number of objects in one frame, categories of extended radio galaxies, and the occupied area of labeled objects, are depicted in Figure~\ref{FIG:Statistics}. 
Additionally, Table~\ref{TAB:DataComparison} provides a comparison with the existing MiraBest \citep{miraghaei17} and Citizen Science RGZ \citep{wu19} datasets.

\begin{table}
  \caption{Existing datasets for radio galaxy classification. The annotations C, B, S, and K are categories, bounding boxes, segmentation and keypoint labels, respectively.}
  \label{sample-table}
  \centering
  \begin{tabular}{lcllcc}
    \toprule \\
    Name                    & \#Complex & Annot.      & Radio Image & Domain  & Image Size\\
                            & Galaxies  & type        & noise ($\mu$Jy/b) & experts      & (pixels)\\
    \midrule
    MiraBest                & 1,256     & C           & $\sim140$  & \cmark & $150\times150$\\
    Citizen Science RGZ     & 6,536     & C, B        & $\sim140$  & \xmark & $132\times132$\\
    Present work        & 2,800      & C, B, S, K  & $\sim30$   & \cmark & $450\times450$\\
    \bottomrule
  \end{tabular}
  \label{TAB:DataComparison}
\end{table}
The process of obtaining annotations for our dataset took nearly 1.5 years, involving multiple discussions over each source, marking the first such dataset in radio astronomy that utilizes such extensive scientific resources. 

\section{Experiments}
\label{SEC:Experiments}
\begin{figure}[!ht]
\centering
\includegraphics[trim={0cm 0cm 7.8cm 0cm}, clip,width=13.8cm, scale=0.5]{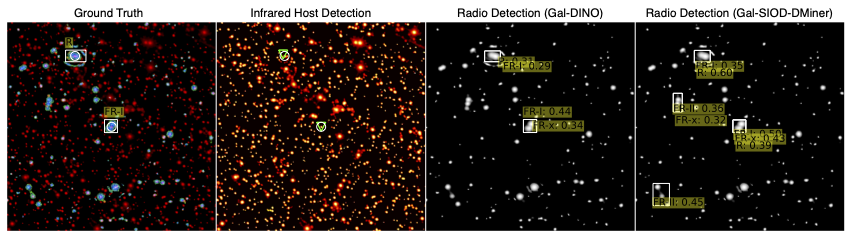}
\includegraphics[trim={0cm 0cm 7.8cm 1.55cm}, clip, width=13.8cm, scale=0.5]{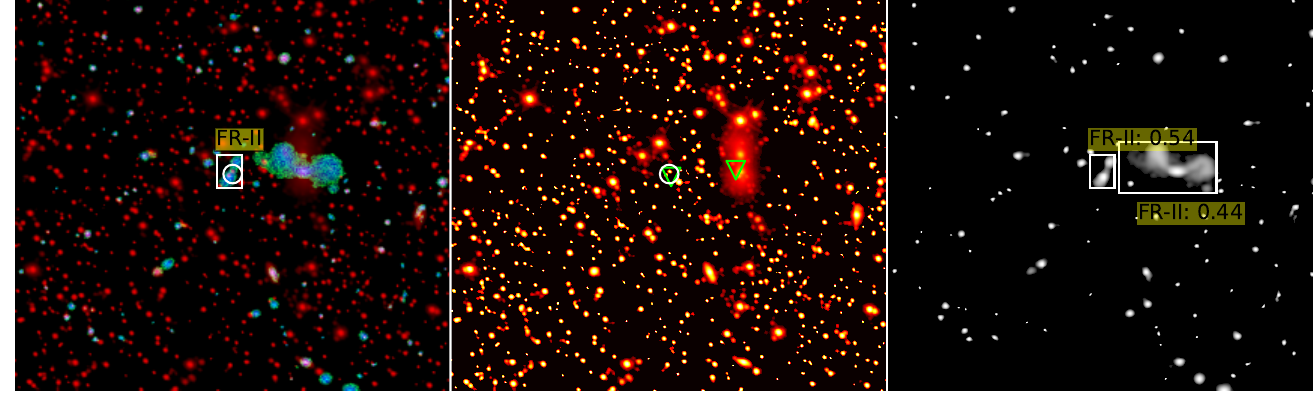}
\includegraphics[trim={0cm 0cm 7.8cm 1.55cm}, clip, width=13.8cm, scale=0.5]{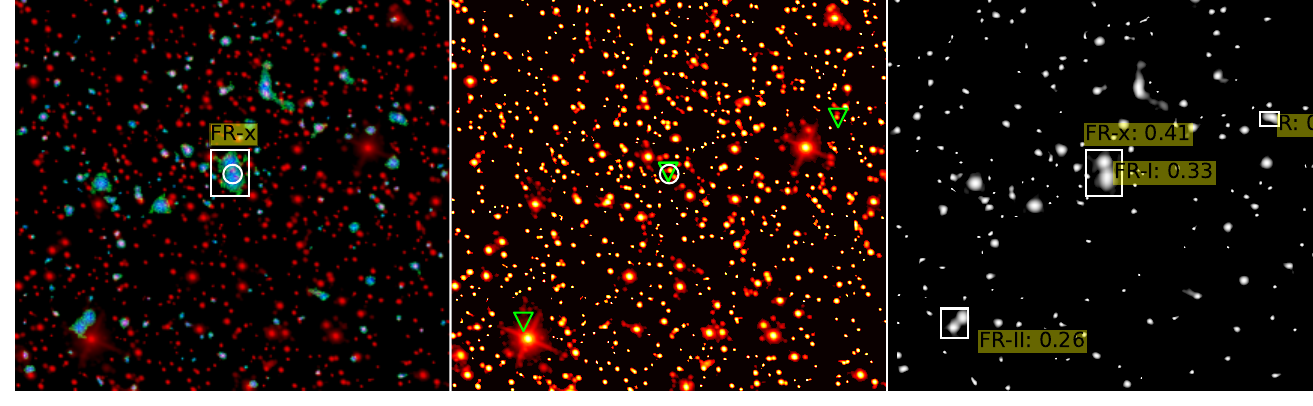}
\caption{Object detection results: Shown are the processed radio-radio-infrared images and ground truth annotations (first column), ground truth and Gal-DINO keypoint detections as circles and triangles over infrared images (second column), Gal-DINO (third column) class and bounding box predictions over radio images.} 
\label{FIG:Results}
\end{figure}

\begin{table}
    \centering
    \caption{Bounding box and keypoint detection results on the test set.}
    \begin{NiceTabular}{clcccccccc}
    \toprule
    &  Model    & Params & Epochs      & AP  & AP$_{50}$ & AP$_{75}$  & AP$_{\rm S}$ & AP$_{\rm M}$ & AP$_{\rm L}$ \\
    &           &        &             & (\%)  & (\%)  & (\%)  & (\%) & (\%) & (\%) \\
    \midrule
    \Block{3-1}{\rotate Bbox}
    & Gal-DETR            & 41M  & 500     & 22.6  & 38.1  & 26.2  & 16.3  &  24.8 & 19.8 \\
    & Gal-Deformable DETR & 40M  & 100     & 40.2  & 52.1  & 45.9  & 37.7  &  39.9 & 22.2 \\
    & Gal-DINO-4scale     & 47M  & 30      & 53.7  & 60.2  & 58.9  & 41.5  &  56.9 & 35.2 \\
    \midrule
    \Block{3-1}{\rotate Keys}
    & Gal-DETR            & 41M  & 500     & 35.4  & 37.5  & 35.3  & 9.1   &  60.0 & 49.6 \\
    & Gal-Deformable DETR & 40M  & 100     & 45.0  & 49.0  & 45.3  & 21.5  &  79.9 & 76.1 \\
    & Gal-DINO-4scale     & 47M  & 30      & 48.1  & 53.4  & 48.4  & 17.6  &  81.4 & 82.9 \\
    \bottomrule
    \end{NiceTabular}
    \label{TAB:AP1}
\end{table}
We propose a novel multimodal modelling approach to simultaneously detect radio galaxies and their corresponding infrared hosts by incorporating keypoint detection in existing object detection algorithms.
Note that the multimodal methods are tailored to specific tasks. Here we have radio images where galaxies appear larger due to extended emission, while in infrared images, the same galaxies look like point objects (as depicted in columns 2 and 3 of Figure~\ref{FIG:RadioGalaxies15arcm}). To the best of our knowledge, there are no specific models that deal with objects that look completely different in two image modalities. As a result, we introduce our own approach to multimodal modelling.

We implemented keypoint detection for Gal-DETR \citep[based on DETR;][]{carion2020end}, Gal-Deformable DETR \citep[based on Deformable DETR][]{zhu2021deformable}, and Gal-DINO \citep[based on DINO;][]{zhang2022dino}.
Specifically, we implemented keypoint detection to the model, augmentations, and Hungarian matcher and added additional random rotation augmentations during training. 
We reduced the learning rate to $5\times 10^{-5}$ and the number of queries to 10. Similar changes were made for Gal-Deformable DETR model, where keypoint detection was also implemented in the deformable attention mechanism. 
For Gal-DINO model, we made the same changes as for Gal-DETR and additionally implemented keypoint detection in the de-noising anchor box mechanism.
All networks are trained and evaluated on an Nvidia Tesla P100.
Table~\ref{TAB:AP1} presents the results of Gal-DETR, Gal-Deformable DETR, and Gal-DINO for bounding box detection of extended radio galaxies and keypoint detection for the positions of infrared host galaxies, evaluated using the COCO evaluation metric.
Figure~\ref{FIG:Results} displays RGB images and ground truth annotations (first column), ground truth and predicted keypoints as circles and triangles over infrared images (second column) and Gal-DINO bounding box predictions over radio images. All predictions are above the confidence threshold of 0.25.
Further details about the dataset and benchmarks are be available in \citet{gupta2023b}.

\section{Conclusions}
We present a multimodal dataset comprising 2,800 images capturing both radio and infrared sky data, with annotations curated by professional astronomers.
The dataset features 4,155 instances of annotations, including class information of extended radio galaxies, bounding boxes encompassing all associated components of each radio galaxy, segmentation masks for radio galaxies, and positions of host galaxies in infrared images.
We benchmark various object detection strategies on the dataset and propose a novel method for simultaneously detecting the extended radio galaxies and the positions of infrared host galaxies. 
The availability of our dataset will facilitate the development of machine-learning methods to detect radio galaxies and infrared hosts in the next generation of radio sky surveys, enabling the creation of efficient multimodal algorithms with a focus on small objects and partial annotations.

\paragraph{Data and Architecture Availability.} Our dataset can be downloaded from \url{https://doi.org/10.25919/btk3-vx79}.
Network architectures for Gal-DETR, Gal-Deformable DETR and Gal-DINO can be cloned from \url{http://hdl.handle.net/102.100.100/602494?index=1}.  This work is accepted for publication in PASA journal,  DOI: \url{https://doi.org/10.1017/pasa.2023.64}.

\section*{Acknowledgements}
The Australian SKA Pathfinder is part of the Australia Telescope National Facility, which is managed by CSIRO.  
The operation of ASKAP is funded by the Australian Government with support from the National Collaborative Research Infrastructure Strategy.  
ASKAP uses the resources of the Pawsey Supercomputing Centre. 
The establishment of ASKAP,  Inyarrimanha Ilgari Bundara, the Murchison Radio-astronomy Observatory and the Pawsey Supercomputing Centre are initiatives of the Australian Government, with support from the Government of Western Australia and the Science and Industry Endowment Fund. 
We acknowledge the Wajarri Yamatji people as the traditional owners of the Observatory site.
NG acknowledges support from CSIRO’s Machine Learning and Artificial Intelligence Future Science Impossible Without You (MLAI FSP IWY) Platform.


\bibliographystyle{apalike}
\bibliography{main}

\end{document}